\documentclass[pdflatex, twocolumn, sn-mathphys]{sn-jnl}

\jyear{2021}%

\theoremstyle{thmstyleone}%
\theoremstyle{thmstyletwo}%

\theoremstyle{thmstylethree}%

\begin{document}

\title[Article Title]{
{Construction Site Safety Monitoring and Excavator Activity Analysis System}}





\author*[1]{\fnm{Sibo} \sur{Zhang}}\email{sibozhang1@gmail.com}

\author[2]{\fnm{Liangjun} \sur{Zhang}}\email{liangjunzhang@baidu.com}


\affil[1]{\orgdiv{Baidu Research}, \orgname{Baidu USA}, \orgaddress{\street{1195 Bordeaux Dr}, \city{Sunnyvale}, \postcode{94089}, \state{CA}, \country{USA}}}




\abstract{
{With the recent advancements in deep learning and computer vision, the AI-powered construction machine such as autonomous excavator has made significant progress. Safety is the most important section in modern construction, where construction machines are more and more automated.} In this paper, we propose a vision-based excavator perception, activity analysis, and safety monitoring system. Our perception system could detect multi-class construction machines and humans in real-time while estimating the poses and actions of the excavator. Then, we present a novel safety monitoring and excavator activity analysis system based on the perception result. To evaluate the performance of our method, we collect a dataset using the Autonomous Excavator System (AES) \cite{zhang2021autonomous} including multi-class of objects in different lighting conditions with human annotations. We also evaluate our method on a benchmark construction dataset. 
{The results showed our YOLO v5 multi-class objects detection model improved inference speed by 8 times (YOLO v5 x-large) to 34 times (YOLO v5 small) compared with Faster R-CNN/ YOLO v3 model \cite{10.22260/ISARC2021/0009}. Furthermore, the accuracy of YOLO v5 models is improved by 2.7\% (YOLO v5 x-large) while model size is reduced by 63.9\% (YOLO v5 x-large) to 93.9\% (YOLO v5 small).} The experimental results show that the proposed action recognition approach outperforms the state-of-the-art approaches on top-1 accuracy by about 5.18\%. 
{The proposed real-time safety monitoring system is not only designed for our Autonomous Excavator System (AES) in solid waste scenes, it can also be applied to general construction scenarios.}
}




\maketitle

\section{Introduction}\label{sec1}

Operating excavators in a real-world environment can be challenging due to extreme conditions, such as multiple fatalities and injuries occur each year during excavations. Safety is one of the main requirements on construction sites. With the advance of deep learning and computer vision technology, Autonomous Excavator System (AES) has made solid progress \cite{zhang2021autonomous}. In the AES system, the excavator is assigned to load the waste disposal material into a designated area. While the system is capable of operating a whole 24-hour day without any human intervention, in this paper, we mainly address the issue of safety, where the excavator could potentially collide with the environment or other construction machines. We propose a camera-based safety monitoring system that detects the excavator poses, the surrounding environment, and other construction machines, and warns of any potential collisions. In addition, based on action recognition algorithm on human activity, we successfully extend the algorithm to excavator actions and use it to develop an excavator productivity analysis system. We note that although developed for the autonomous excavator, this system can also be generally applied to manned excavators. 

To build an excavator safety monitor system, we first need to build a perception system for the surrounding environment. The perception system includes detection, pose estimation, and activity recognition of construction machines. Detecting the excavator pose in real-time is a key requirement to inform the workers and to enable autonomous operation. Vision-based (marker-less, marker-based) and sensor-based (IMU, UWB) are two of the main methods for estimating robot pose. The marker-based and sensor-based methods require some additional pre-installed sensors or markers, whereas the marker-less method only requires an on-site camera system, which is common on modern construction sites. Therefore, we adopt a marker-less approach and develop the system solely from camera video input, leveraging state-of-the-art deep learning methods. 

In this paper, we propose a deep learning-based excavator activity analysis and safety monitor system which can detect the surrounding environment, estimate poses, and recognize actions of excavators. The main contributions of this paper are summarized as follows:

1) A deep learning-based perception system for real-time multi-class object detection, pose estimation, and action recognition of construction machinery and human on construction sites. 

2) Novel excavator safety monitoring and activity analysis system based on the perception system. 

3) We collect a dataset from the Autonomous Excavator System (AES) including multi-class of objects in different lighting conditions with human annotations. Our network gets SOTA results on the AES dataset and UIUC construction dataset. 



\section{Related Works}\label{sec2}

\begin{figure*}
  \centering
  \includegraphics[width=0.99\textwidth]{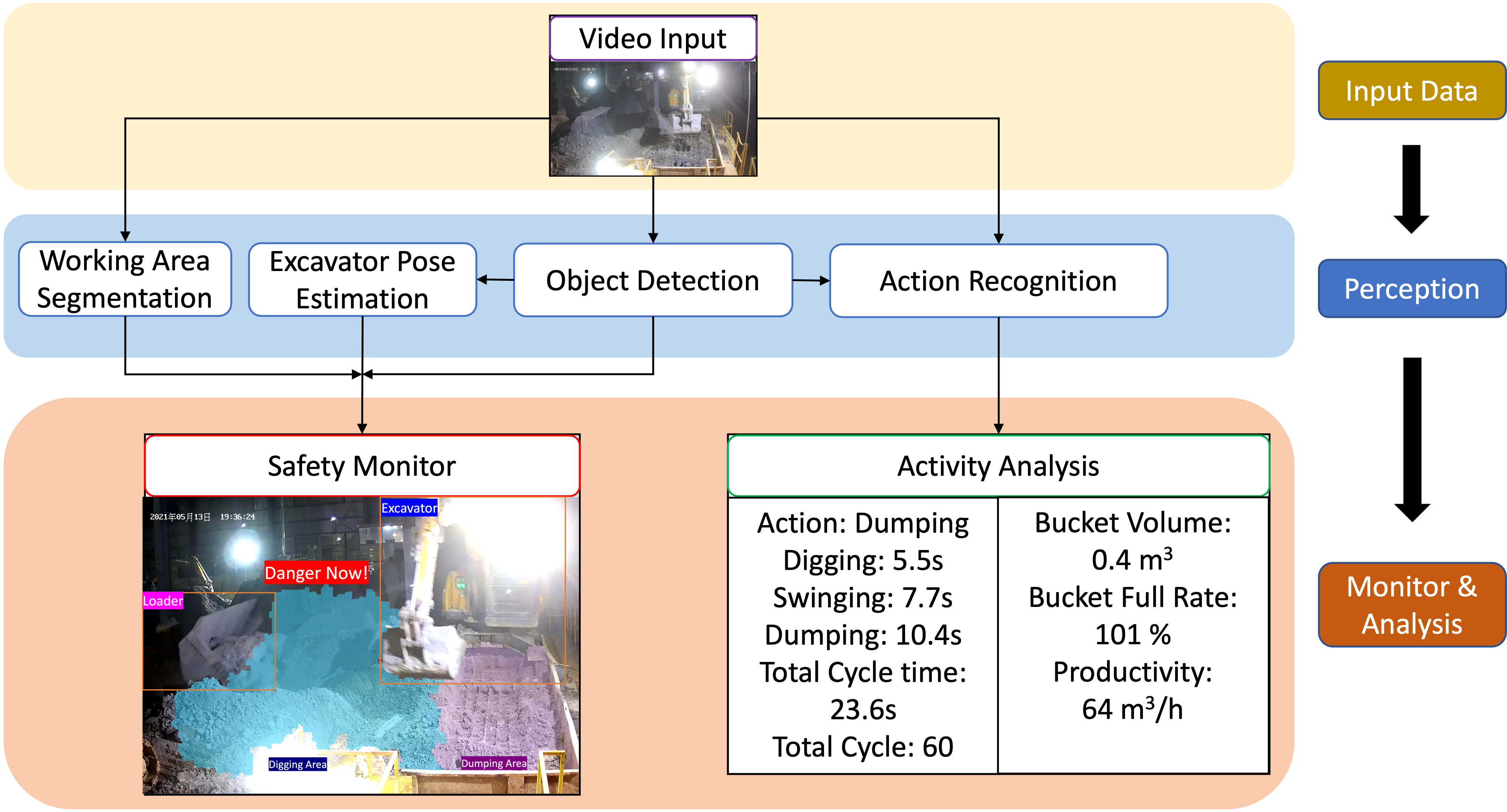}
  \caption{The autonomous excavator activity analysis and safety monitoring system pipeline, including Object Detection, Excavator Pose Estimation, Working Area  Segmentation, Action Recognition, Safety Monitor and Activity Analysis modules. In the Activity Analysis module, the example showing our system could recognize current excavator action, calculate time duration of each actions, total cycle time, total cycles and excavator productivity.}
  \label{fig:pipeline}
\end{figure*}


Previous studies related to safety and productivity analysis are reviewed here. We start with some of the most basic tasks in computer vision that are essential to activity analysis and safety monitoring system, including object detection, image segmentation, pose estimation and action recognition. Then, we review vision-based activity analysis and safety monitoring system.

\textbf{Object Detection.}
The first category is object detection. 
Recently, Wang et al. \cite{wang2019predicting} used a region-based CNN framework named Faster R-CNN \cite{ren2016faster} to detect workers standing on scaffolds. A deep CNN then classified whether workers are wearing safety belts. Those without safety belts appropriately harnessed were identified to prevent any fall from height. 
{ YOLOv3 \cite{redmon2018yolov3} is a one-stage state-of-art detector with extremely fast speed. 
With the original authors work on YOLO coming to a standstill, YOLOv4 was released by Alexey Bochoknovskiy \cite{bochkovskiy2020yolov4}.
Shortly after the release of YOLOv4, Glenn Jocher introduced YOLOv5 \cite{glenn_jocher_2020_4154370} using the Pytorch framework.
YOLOv4 and YOLOv5 utilized Cross Stage Partial Network (CSPNet)  \cite{wang2020cspnet} as backbone. It is designed, to attribute the problem to the duplicate gradient information within network optimization, complexity can be largely reduced while maintaining the accuracy.
}

\textbf{Image Segmentation.}
Raoofi et al. \cite{raoofiamask} used Mask R-CNN to detect construction machinery on Job sites. More importantly, a segmentation network like Mask R-CNN can be used to decide areas like digging and dumping. 

\textbf{Pose Estimation.}
The second group of technology is skeleton pose estimation. Pose estimation has been studied \cite{nakamura2020pose} based on human pose estimation network like OpenPose. Soltani et al. \cite{soltani2017skeleton} proposed skeleton parts estimation of excavators.  

\textbf{Action Recognition.}
Learning-based action recognition methods. Feichtenhofer et al. \cite{feichtenhofer2019slowfast} proposed a SlowFast network for video recognition. The model involves a low pathway that operating at a low frame rate, to capture spatial semantics, and a Fast pathway that operating at a high frame rate, to capture motion at fine temporal resolution. Bertasius et al. \cite{bertasius2021space} presented a convolution-free approach to video classification built exclusively on self-attention over space and time.

\textbf{Activity Analysis and Safety Monitoring.}
Here we review recent vision based activity analysis and safety monitoring methods in the construction area. For example, Ding et al.  \cite{ding2018deep} combined CNN with Long-Short-Term-Memory (LSTM) to identify unsafe actions of workers, such as climbing ladders with hand-carry objects, backward-facing, or reaching far. While safety hazards of workers were effectively identified, their method only captured a single worker, and multi-object analysis was not considered. 
On the other hand, Soltani et al. \cite{soltani2017skeleton} used background subtraction to estimate the posture of an excavator by individually detecting each of its three skeleton parts including the excavator dipper, boom, and body. 
Although knowing the operating state of construction equipment would allow safety monitoring nearby, the influence of the equipment on the surrounding objects was not studied.
Chen et al. \cite{chen2020automated} propose a framework to automatically recognize activities and analyze the productivity of multiple excavators.
Wang et al. \cite{wang2019predicting} proposed a methodology to monitor and analyze the interaction between workers and equipment by detecting their locations and trajectories and identifying the danger zones using computer vision and deep learning techniques. However, the excavator state is not considered in their model. 
Roberts et al. \cite{roberts2019end} proposed a benchmark dataset. However, their action recognition model accuracy is low compared to our deep learning-based model. 

Overall, in terms of activity analysis and safety monitoring with computer vision techniques, previous studies focused on different parts separately, such as identifying the working status of construction equipment or pose estimation of the excavator. Our method combine the advantages of SOTA deep learning models from detection, pose estimation, and action recognition tasks.

\section{Proposed Framework}\label{sec3}

The perception framework is shown in Fig.~\ref{fig:pipeline}. The framework contains six main modules: Object Detection, Excavator Pose Estimation, Working Area Segmentation, Action Recognition, Safety Monitor and Activity Analysis. The input to our system is surveillance camera video. First, working areas are being segmented into digging and dumping areas. Then, the detection method is used to identify all construction machines in video frames with equipment type. Second, the excavator is identified through pose estimation and detection-based tracking.
Then, the action state of the tracked excavators is recognized with pose estimation and working area segmentation. Finally, construction site safety is monitored based on detection and activity recognition results. Besides, the productivity of the excavator is calculated by the activity recognition results. The details about each module in the framework are provided in the following sub-sections.

\subsection{
{Multi-class Construction Machines and Human Detection} }

Faster R-CNN \cite{ren2016faster}. The architecture of Faster R-CNN includes (1) backbone network to extract image features; (2) region proposal generate (RPN) network for generating region of interest (ROI), and (3) classification network for producing class scores and bounding boxes for objects. To remove duplicate bounding box, we applied Soft-NMS \cite{bodla2017soft} to limit max bounding box per object to 1.

{
To improve inference speed, reduce model size and further improve detection accuracy, we implement our real-time detection of construction machines and human network based on YOLOv5. YOLOv5 has four different size models including YOLOv5s,
YOLOv5m, YOLOv5l and YOLOv5x. Generally, YOLOv5 respectively uses the architecture of CSPDarknet53 with an SPP layer as backbone, PANet as Neck and YOLO detection head \cite{redmon2016you}. To further optimize the whole architecture,
bag of freebies and specials \cite{bochkovskiy2020yolov4} are provided. Since it is the most notable and convenient one-stage detector, we select it
as our baseline. To improve human detection accuracy in all scenery, we fine-tune pretrained YOLOv5 model on our construction dataset.
} 

\subsection{Excavator Pose Estimation}
\begin{figure}
  \centering
  \includegraphics[width=0.45\textwidth]{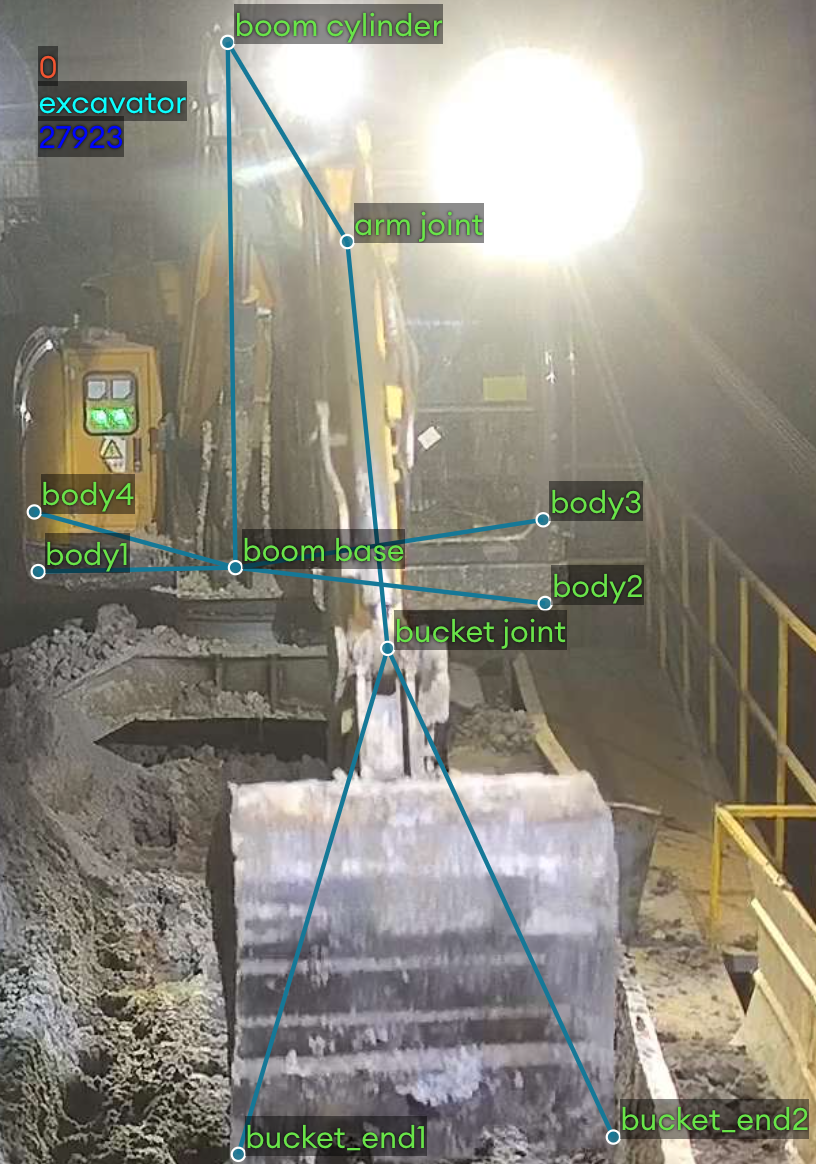}
  \caption{Excavator and corresponding pose labels. We labeled 10 parts of excavators including 2 bucket end keypoints (bucket end1, bucket end2), bucket joint, arm joint, boom cylinder, boom base and 4 body keypoints (body1, body2, body3, body4).}
  \label{fig:excavator_pose_label}
\end{figure}

The pose estimation is based on the output bounding box from detection. We use SimpleBaseline ~\cite{xiao2018simple} for pose estimation. The network's backbone is ResNet. We design a labeling method for the fixed crawler excavator as 10 keypoints. Those keypoints of excavator parts annotated are shown in Fig.~\ref{fig:excavator_pose_label}. These 10 keypoints including 2 bucket end keypoints, bucket joint, arm joint, boom cylinder, boom base, and 4 body keypoints. Unlike other pose label methods ~\cite{nakamura2020pose} to label bucket/ excavator body as the middle point, we label corner point to improve accuracy.

\subsection{Working Area Segmentation}
\begin{figure}
  \centering
  \includegraphics[width=0.49\textwidth]{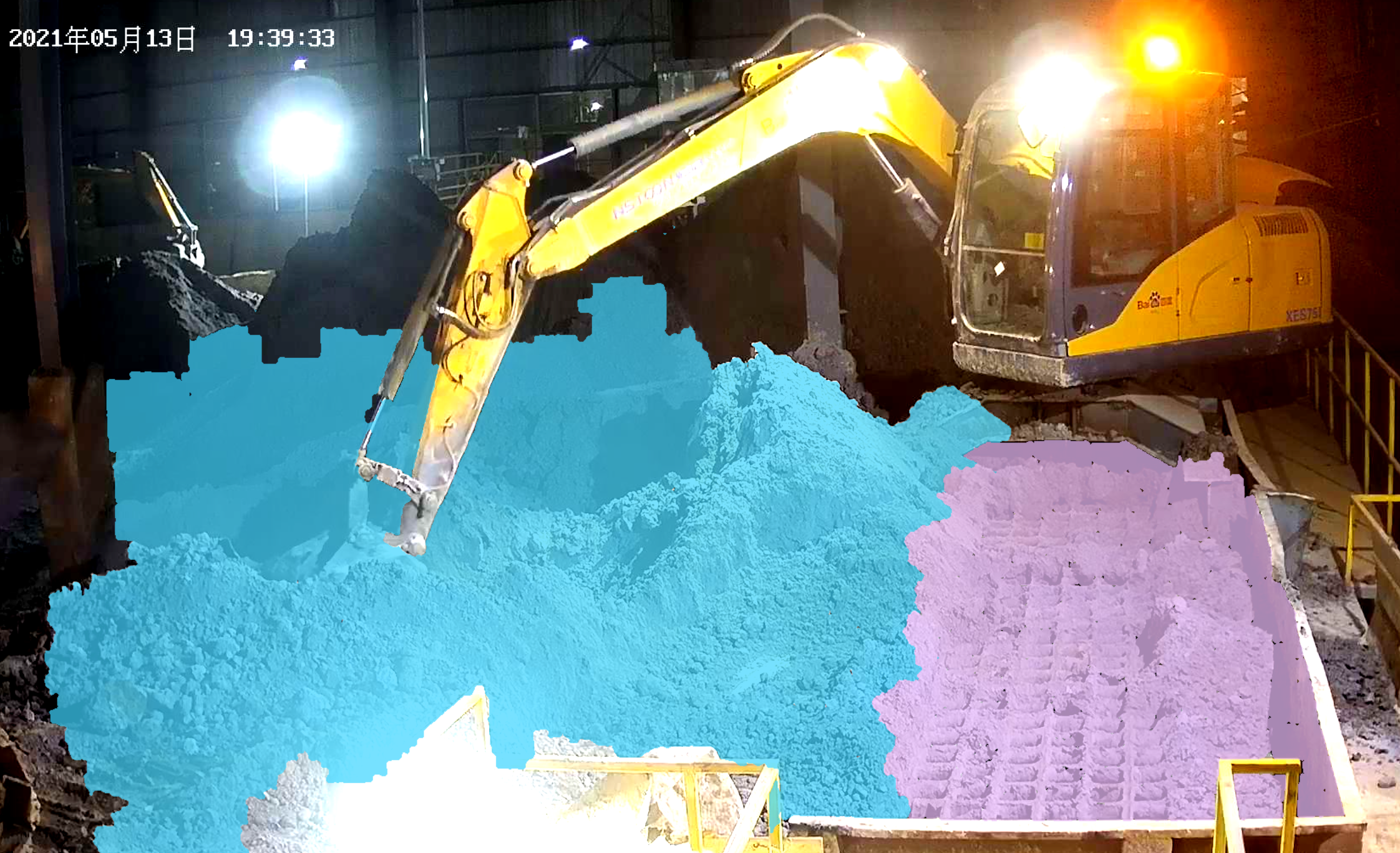}
  \caption{Area segmentation. The pink color area is dumping area and the blue color area is digging area.}
  \label{fig:area_segmentation}
\end{figure}

We use image segmentation to decide digging and dumping areas as shown in Fig.~\ref{fig:area_segmentation}. The segmentation network is based on ResNet \cite{he2016deep}. A digging area is defined as the waste recycling area which including various toxic materials. A dumping area is a designated area to dump waste.

\subsection{Excavator Action Recognition}

We define three actions for excavator: 1. Digging 2. Swinging 3. Dumping. Specifically, we define four states of our autonomous excavator: 1. Digging state 2. Swinging after digging state 3. Dumping state 4. Swinging for digging state. More precisely, Digging indicates loading the excavator bucket with target material; Swinging after digging indicates swinging the excavator bucket to the dumping area; Dumping means unloading the material from the bucket to the dumping area, and Swinging for digging means swinging the bucket to the working area. Besides, there is an optional idle state when the excavator is in manned mode or malfunction status. 


To determine the excavator action state, we first determine excavator position based on keypoints from pose estimation and image segmentation results.
Then we use continuous frames of pose keypoints of body1 to body4 as shown in Fig.~\ref{fig:excavator_pose_label} to decide whether the excavator is in the swing state. We set a threshold for keypoints movement: if the mean of pose keypoints of body1 to body4 movements is smaller than a set value, then we think the excavator body is still. Otherwise, we think the excavator body is not still. This rule-based module is used in our safety monitor system. Our excavator action states are defined as follows:

1. Digging state: buckets/ arm joint in digging area and body1 to body4 are fixed points (excavator body is stilled).

2. Swinging state: buckets/ arm joint in working area and body1 to body4 are not fixed points (excavator body is not stilled). Then we can decide whether it is Swing for digging state or Swing after digging state by the previous state. If the previous state is a Dumping state then it will be Swing for digging state. Otherwise, it will be Swing after digging state.

3. Dumping state: buckets/ arm joint in dumping area and body1 to body4 are fixed points (excavator body is stilled).

4. Idle state: buckets/ arm joint in dumping area and buckets/ arm joint/ body1 to body4 are fixed points (excavator arm and body are both stilled).

Then, we implement a more general deep learning-based action recognition method based on SlowFast \cite{feichtenhofer2019slowfast}. The input to the network is the surveillance video. The model involves (i) a Slow pathway, operating at a low frame rate, to capture spatial semantics, and (ii) a Fast pathway, operating at a high frame rate, to capture motion at fine temporal resolution. The Fast pathway can be made very lightweight by reducing its channel capacity, yet can learn useful temporal information for video recognition. This deep learning action recognition model is used in the Activity Analysis module.


\subsection{Safety Monitor}

\begin{figure*}
  \centering
  \includegraphics[width=0.99\textwidth]{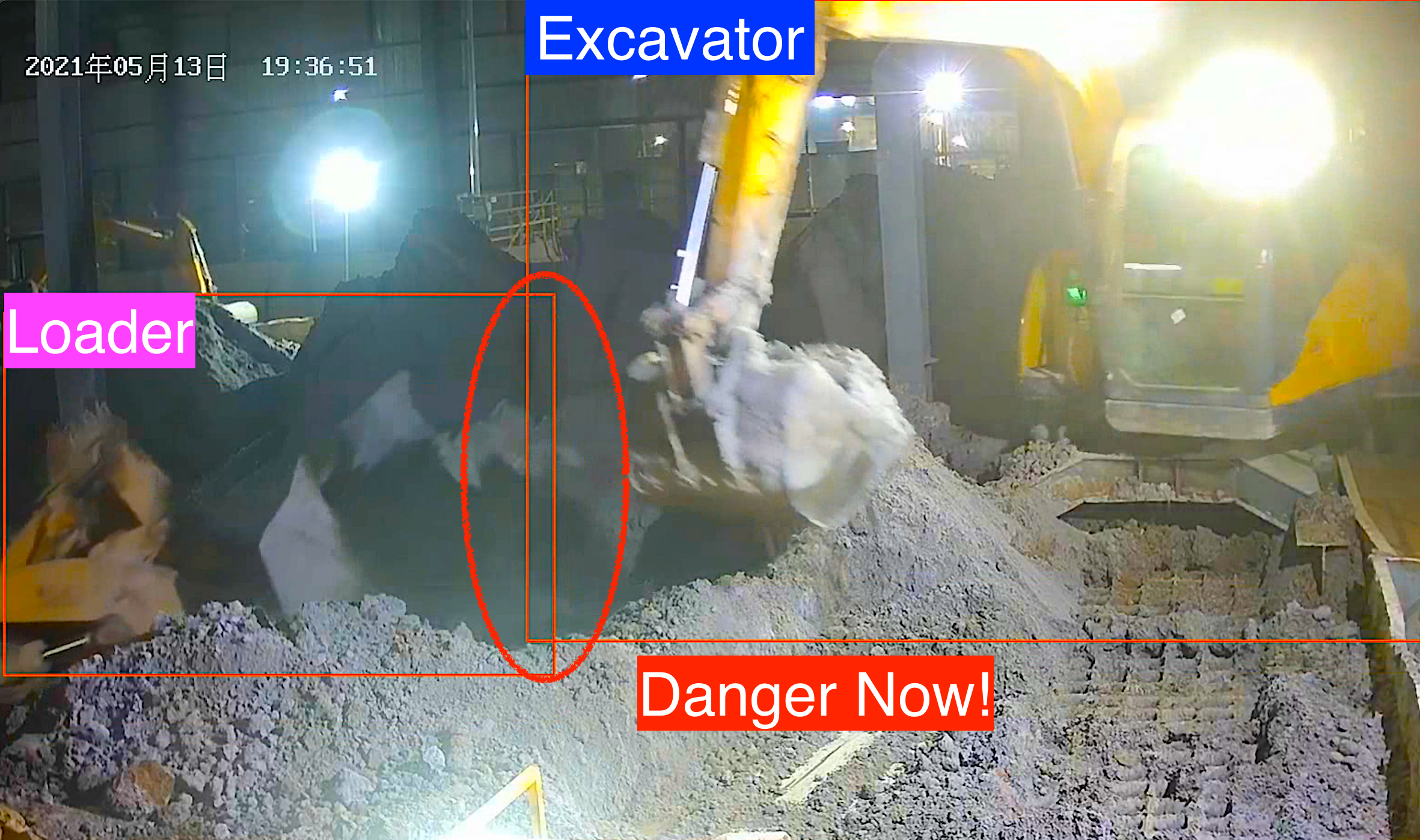}
  \caption{The autonomous excavator and loader potential collision scene when loader tries to load in digging area. The danger signal is sent when the autonomous excavator and the loader machines are both detected in the digging area.
  }
  \label{fig:excavators_loader_potential_collision}
\end{figure*}

\textbf{Detect Potential Construction Machine Collision.} 
The autonomous excavator and the loader may have potential collision as Fig.~\ref{fig:excavators_loader_potential_collision} shows. So it is important to detect potential collision since the loader is hard to know which state excavator is currently at from his view. 
If more than one machine is detected with in the same region (digging or working area), then an alert will be indicated to the user, and the autonomous vehicles will pause until the issue is cleared. 


\subsection{Productivity Analysis}


\begin{figure*}
  \centering
  \includegraphics[width=0.99\textwidth]{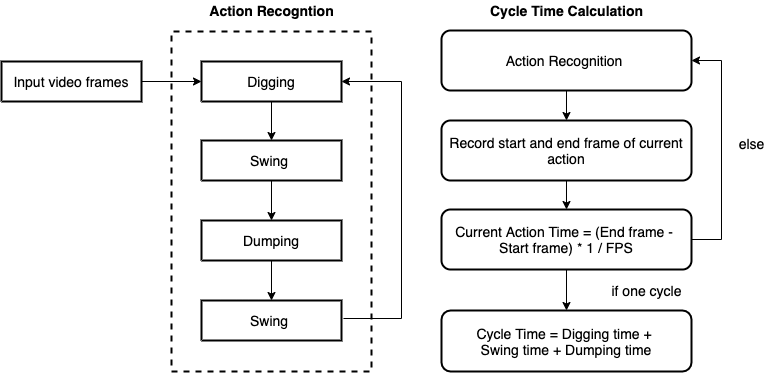}
  \caption{Excavator working cycle.}
  \label{fig:excavator_working_cycle}
\end{figure*}

The productivity of the excavator is based on the activity recognition results. In the solid waste recycle scene, excavators usually work with other equipment, such as loaders. For example, an excavator digs the waste and dumps it into a dumping area. When waste is empty in the digging area, the loader will load and dump waste in the digging area. The excavator's productivity can be calculated with the cycle time, the bucket payload and the average bucket full rate, as shown in Equation \ref{eq_prductivity}. Since the bucket payload is given by the manufacturer, the target of the productivity calculation becomes to determine the cycle time of the excavator and the bucket full rate. To simplify the procedure, the two types of swinging (swinging after digging and swinging for digging) are not distinguished in this paper. 

\begin{equation}
\label{eq_prductivity}
\begin{split}
    &Productivity (m^3/hr) = \frac{Cycles}{hr} \times \\
    & Bucket Volume (m^3) \times Bucket Full Rate
\end{split}
\end{equation}

The time for each cycle is measured by the workflow showing in Fig.~\ref{fig:excavator_working_cycle}. Our action recognition module labels each video frame of the excavator with an action label. Next, the action labels of two consecutive frames are compared. If they are the same, it means that the action remains same. Thus, the cumulative time for the current action is increased by 1/FPS (frame per second). If the labels are different, it means that a new action has started, and the time of the newly recognized activity will increase by 1/FPS. We define the total time of one cycle as the difference between the start times of two neighboring digging actions.

\section{Experiments}\label{sec4}

\subsection{Dataset}
{We collect an excavator dataset from our Autonomous Excavator System (AES) from the waste disposal recycle scene \cite{zhang2021autonomous}. The dataset including 10 hours of videos containing 9 classes of objects (excavators, loaders, human, truck, crane, cone, hook, car, shovel) in 5 data scenes (AES-line1, AES-line2, bird eye view construction sites, crane construction site, cones dataset). The dataset have 6692 images with object detection bounding boxes, 601 images with excavator poses, and background segmentation.} 
80\% of the images are used for model training while 20\% are for model validation and testing. Besides, we labeled 102 clips of excavator videos with 3 actions (digging, dumping, swinging). The videos were captured at 1920*1080 and filmed at 25 frames per second.

We also test our method based on the benchmark UIUC dataset \cite{roberts2019end} which including 479 action videos of interacting pairs of excavators and dump trucks performing earth-moving operations, accompanied with annotations for object detection, object tracking, and actions. The videos were captured at 480*720 and filmed at 25 FPS.

\subsection{Evaluation}

\subsubsection{Object Detection Evaluation}
The detection evaluation metrics are based on the Microsoft COCO dataset \cite{lin2014microsoft}. The network’s performance is evaluated using Average precision (AP). Precision measures how many of the predictions that the model made were correct and recall measures how well the model finds all the positives. For a specific value of Intersection over Union (IoU), the AP measures the precision/recall curve at recall values (r1, r2, etc.) when the maximum precision value drops.  The AP is then computed as the area under the curve by numerical integration. The mean average precision is the average of AP in each object class. More precisely, AP is defined as:
\begin{equation}
    AP =\frac{1}{11} \sum_{r\in \{0.0,0.1,\dots,1\}} AP_r ,
	\label{equ:AP}
\end{equation} 

\subsubsection{Pose Estimation Evaluation}
The pose estimation matrix is based on the COCO evaluation, which defines the object keypoint similarity (OKS). It uses the mean average precision (AP) over the number of classes for OKS thresholds as main competition metric. The OKS is calculated from the distance between predicted points and ground truth points of the construction machine.

\subsubsection{Action Recognition Evaluation}
The performance metric is the mean Average Precision (mAP) over each object class, using a frame-level IoU threshold of 0.5.

\subsection{Accuracy}
\subsubsection{Accuracy of the detection model}
We implement experiments on the Faster R-CNN model with a backbone network of Resnet-50-FPN and Resnet-152-FPN. The model achieved high detection accuracy for construction equipment. The Average Precision (AP) values of the excavator achieved 93.0\% and the loader achieved 85.2\%. With an mAP of 90.1\%, the model is demonstrated to be promising for detecting multi-class construction equipment accurately on the construction site.


\begin{figure*}
  \centering
  \includegraphics[clip,width=\textwidth]{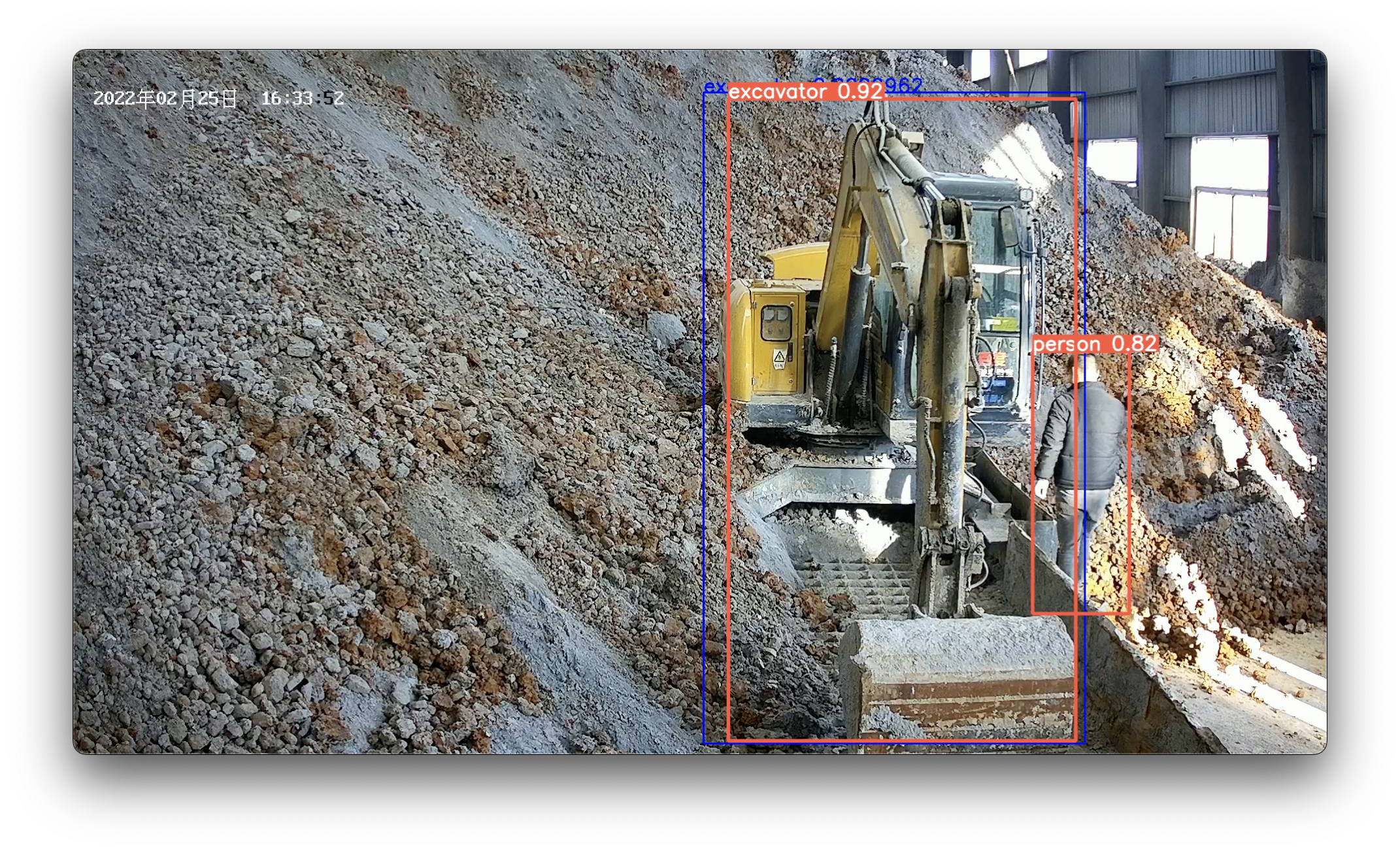}

  \includegraphics[clip,width=\textwidth]{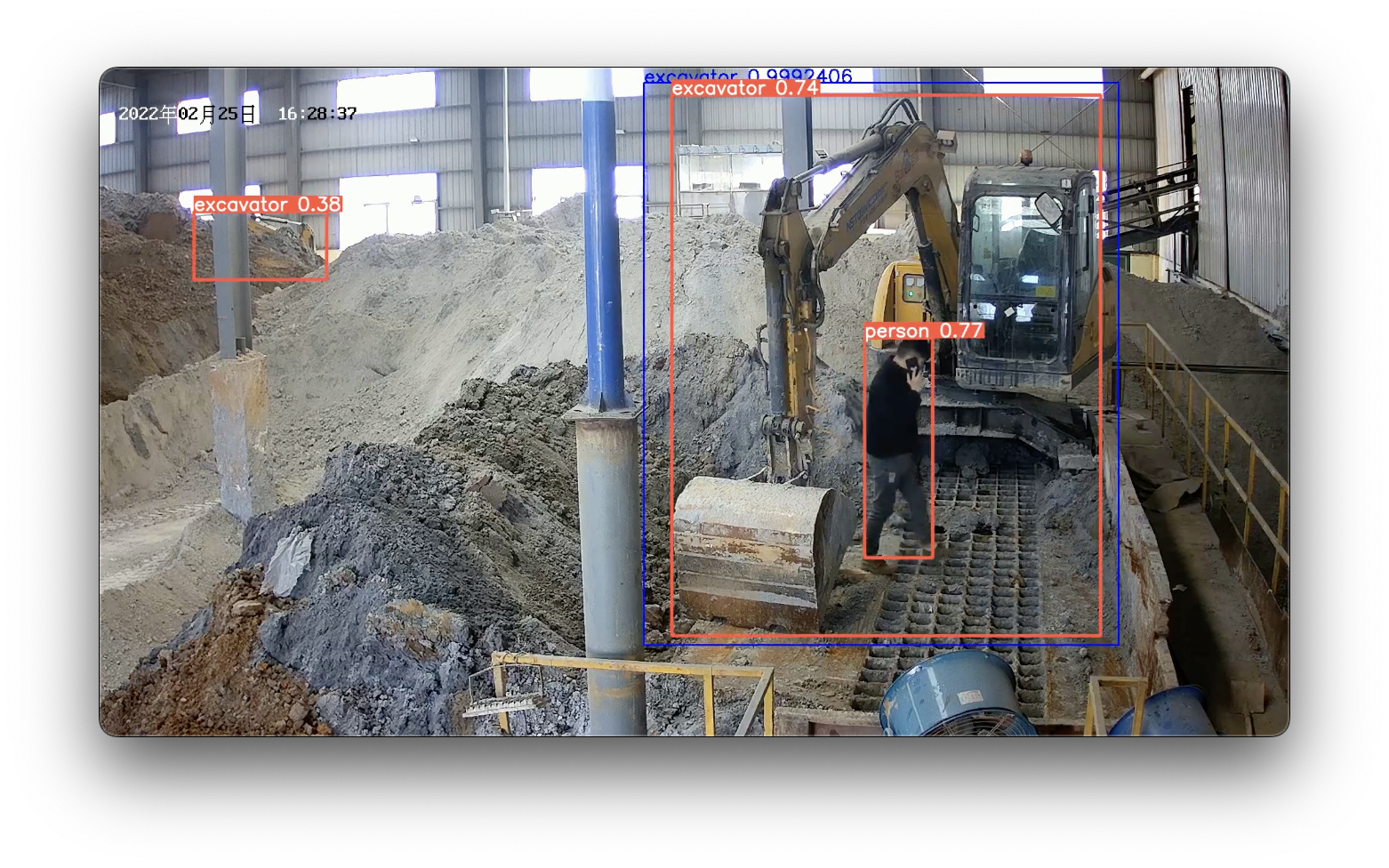}
  {
  \caption{Detection result comparison (Upper figure is from AES-line1 while lower figure from AES-line2). Our perception system fine-tuned on yolov5 pretrained model (orange box) is capable of detecting hard-to-observe human compared to faster-rcnn model (blue box) which missed the human in both scenarios.}
  }
  \label{fig:detection}
\end{figure*}


\begin{table*}[]
\centering
\caption{Accuracy of multi-class construction machine and human detection.}
\label{tab:detection}
\resizebox{.95\textwidth}{!}{
\begin{tabular}{@{}llccc@{}}
\toprule
Network      & Backbone(scale)  & mAP$\!$ (\%)   & Inf. time(ms/frame) & Model Size(MB) \\ \midrule
Faster R-CNN & Resnet-50-FPN     & 90.1       & 588           & 482            \\
Faster R-CNN & Resnet-101-FPN    & 92.3       & 588           & 482            \\
YOLOv3       & DarkNet-53 (320)  & 78.0       & 313           & 492            \\
YOLOv3       & DarkNet-53 (608)  & 75.7       & 344           & 492            \\

YOLOv5s       & CSP-Darknet53 (640) & 88.9       & 9.0           & 14.9             \\
YOLOv5m       & CSP-Darknet53 (640) & 93.0       & 71.4           & 42.9             \\
YOLOv5x       & CSP-Darknet53 (640) & 95.0       & 25.6           & 174.2             \\
\bottomrule
\end{tabular}
}
\end{table*}


We also compared the result with YOLOv3 \cite{redmon2018yolov3}. Compared with some two-stage detectors like Faster R-CNN, the performance of YOLOv3 is slightly low, but the speed is much faster and that is important for real-time applications. We measure the speed of network in ms per frame.

{To further improve model speed and detection accuracy (especially on human), we implement experiments on the YOLO v5 model (small/ medium/ x-large). 
The results showed our YOLO v5 multi-class objects detection model improved inference speed by 8 times (YOLO v5 x-large) to 34 times (YOLO v5 small) compared with Faster R-CNN/ YOLO v3 model \cite{10.22260/ISARC2021/0009}. Furthermore, the accuracy of YOLO v5 models is improved by 2.7\% (YOLO v5 x-large) while model size is reduced by 63.9\% (YOLO v5 x-large) to 93.9\% (YOLO v5 small).}
The detailed comparison result is shown in Table~\ref{tab:detection}. 
The YOLO v5 results are shown in Fig.~\ref{fig:detection}.

\subsubsection{Accuracy of the Pose Estimation}

\begin{figure*}
  \centering
  \includegraphics[width=0.99\textwidth]{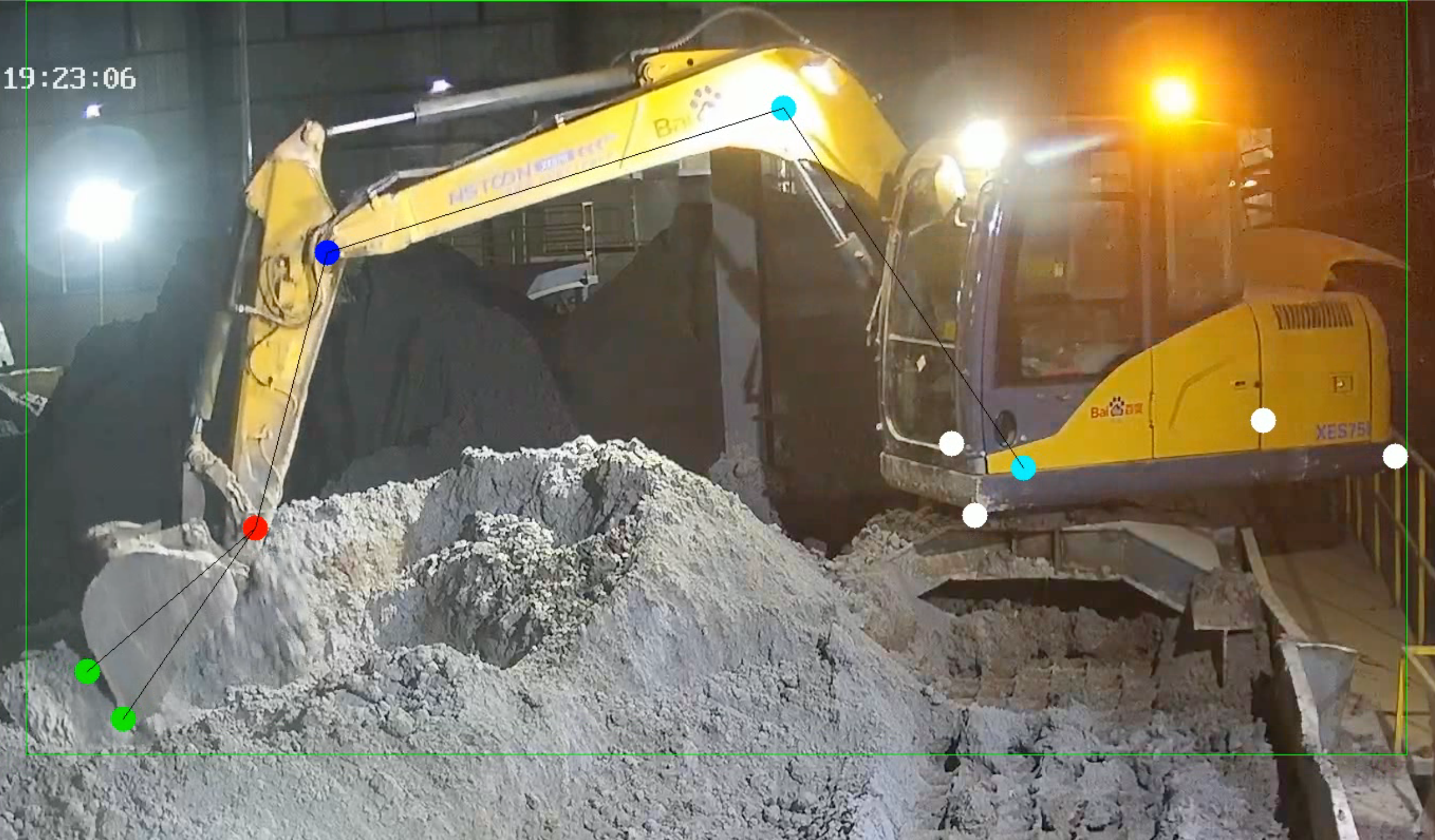}
  \caption{Excavator pose estimation result.}
  \label{fig:excavators_pose}
\end{figure*}

We apply SimpleBaseline~\cite{xiao2018simple} to our pose estimation model and get the following result. Experiments have been conducted on different Backbone networks including Resnet-50 and Resnet-152. Besides, experiments on different image input sizes have been implemented.
The detailed comparison result is shown in Table~\ref{tab:pose_estimation}. The result is shown in Fig.~\ref{fig:excavators_pose}.

\begin{table}[]
\centering
\caption{Accuracy of the SimpleBaseline pose estimation model.}
\label{tab:pose_estimation}
\begin{tabular}{@{}llll@{}}
\toprule
Backbone   & Input size & AP (\%) \\ \midrule
Resnet-50  & 256*192    & 91.79   \\
Resnet-50  & 384*288    & 94.19   \\
Resnet-152 & 384*288    & 96.50   \\ \bottomrule
\end{tabular}
\end{table}



\begin{figure*}
  \centering
  \includegraphics[width=0.99\textwidth]{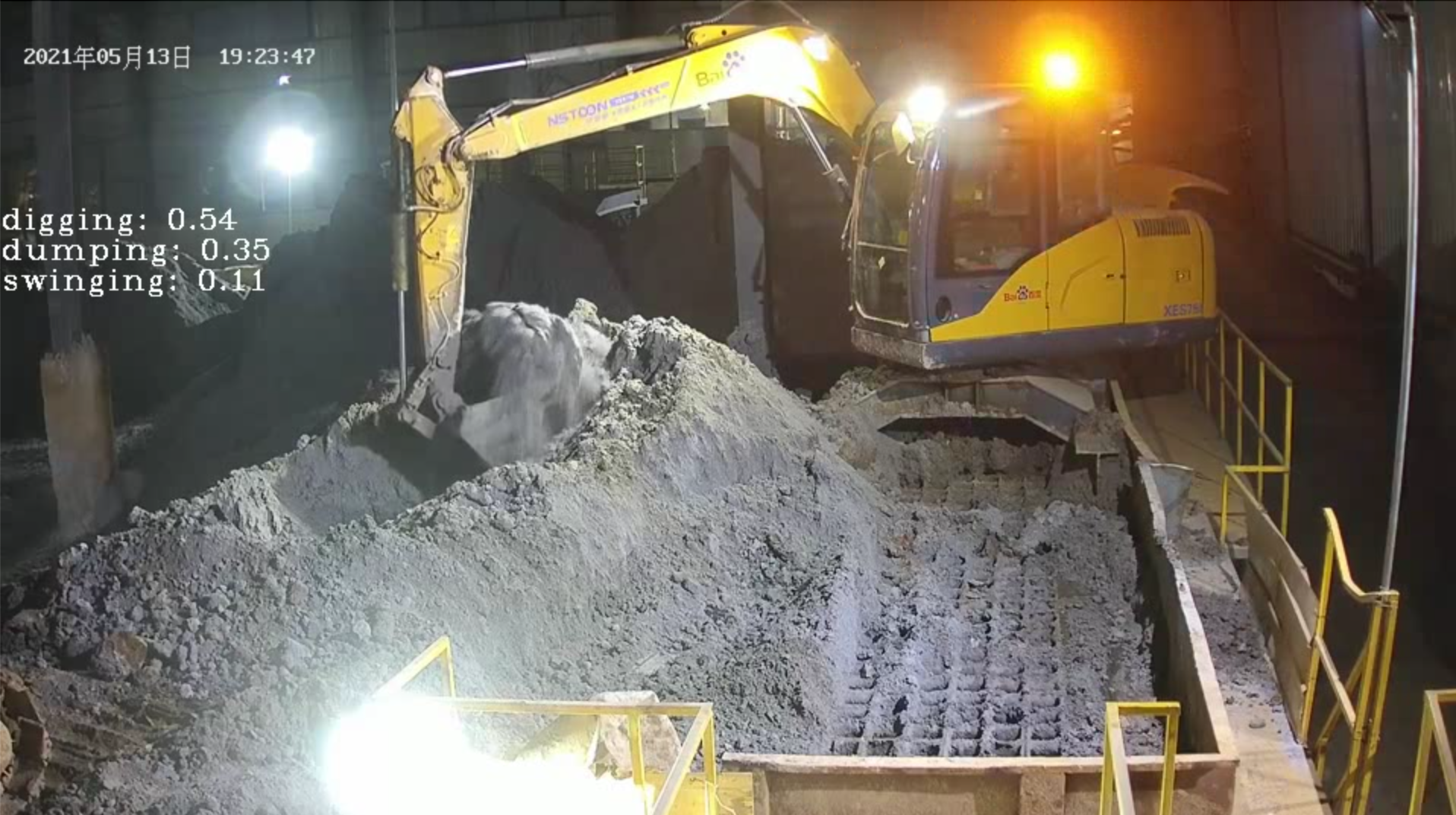}
  \caption{Excavators long video action detection result. Action recognition with the highest possibility is showing in the first line.}
  \label{fig:digging_result}
\end{figure*}

\begin{table*}[]
\centering
\caption{Accuracy of the action recognition model on the AES dataset and UIUC dataset from \cite{roberts2019end}.}
\label{tab:action_recognition_accuracy}
\begin{tabular}{@{}llll@{}}
\toprule
Dataset                 & Network                   & Backbone & Top1 Acc. (\%) \\ \midrule
\multirow{2}{*}{AES}    & SlowFast-50               & ResNet3d & 89.70          \\
                        & SlowFast-152              & ResNet3d & 91.44          \\ \midrule
\multirow{3}{*}{UIUC}   & Roberts\cite{roberts2019end} & N/A      & 86.8        \\
                        & SlowFast-50               & ResNet3d & 91.9           \\
                        & SlowFast-152              & ResNet3d & 93.3           \\ 
\bottomrule                              
\end{tabular}
\end{table*}


\begin{figure*}
  \centering
  \includegraphics[width=1\textwidth]{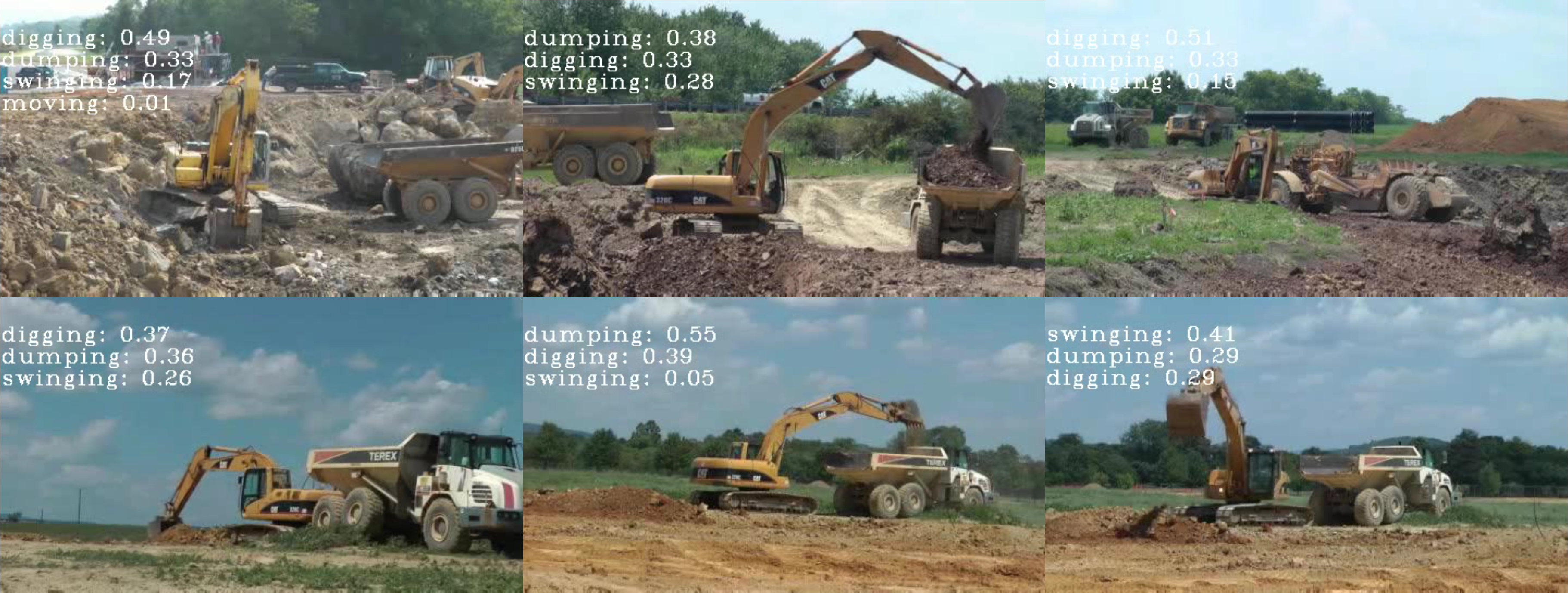}
  \caption{Long video demos of action recognition result on different scenes of the construction dataset. Action recognition with the highest possibility is showing in the first line.}
  \label{fig:action_recognition_result}
\end{figure*}

\subsubsection{Accuracy of the Action Recognition}

We applied Slow-Fast~\cite{feichtenhofer2019slowfast} to our action recognition model and get the following result. Experiments have been conducted on the different networks including SlowFast-101 and SlowFast-152. Besides, experiments on different clip lengths have been implemented.
The detailed comparison result is shown in Table~\ref{tab:action_recognition_accuracy}. The result of top 3 action recognition is showing in the Fig.~\ref{fig:digging_result}. We input a excavator video and the system can predict action result in almost real-time. Action recognition with the highest possibility is showing in the first line. Here the system predict the action as digging with 54\% confidence.

Comparing our result with Roberts~\cite{roberts2019end} on their UIUC dataset, our proposed action recognition approach outperforms their accuracy by about 5.18\%. The action recognition video demo result of the construction dataset is showing in Fig.~\ref{fig:action_recognition_result}. The result shows the advantage of using deep learning model on action recognition task over their Hidden Markov Model (HMM) + Gaussian Mixture Model (GMM) + Support Vector Machine (SVM) method.


\subsection{Activity Analysis}
The proposed framework was tested to estimate the productivity of excavators on a long video sequence, which contains 15 min of excavator's operation. In our video, the XCMG 7.5-ton compact excavator (bucket volume of 0.4 $m^3$) completed 40 working cycles in 15 minutes. The average bucket full rate is 101\% based on our human measurement. So the excavation productivity is 64.64 $m^3$/h according to Equation \ref{eq_prductivity}. Our system detects 39 working cycles in the video which the accuracy of productivity calculation is 97.5\%. The test results showed the feasibility of using our pipeline to analyze real construction projects and to monitor the operation of excavators.

\subsection{Implementation Details}

We implement our detection module based on \textcolor{blue}{YOLOv5 by ultralytics}, MMDetection, segmentation module based on MMSegmentation, pose estimation module based on MMPose, and action recognition module based on MMAction2 toolbox \cite{glenn_jocher_2020_4154370, chen2019mmdetection, mmseg2020, mmpose2020,2020mmaction2}. We use NVIDIA M40 24GB GPUs to train the network. We test on local NVIDIA 1080 GPU. Then we implement optimized version on remote solid waste scene computer with Intel 9700 CPU (16 GB) and NVIDIA 1660 GPU (16 GB).

\subsection{Training and Inference Time}

{ It takes 2, 3, 4 hours to train YOLOv5 small, medium, extra-large model for detection respectively, while 6 hour for pose estimation, and action recognition modules.
The inference time on Nvidia m40 machine for YOLOv5 small detection network can achieve as fast as 9 ms/frame, medium can 14 ms/frame, while extra-large could achieve 39 ms/frame as shown in Table~\ref{tab:detection}. }

\section{Conclusion}\label{sec5}

In this study, we proposed a safety monitor and activity analysis system based on computer vision and deep learning techniques. We integrate object detection, pose estimation, activity recognition modules into our system. Besides, we collect a benchmark dataset including multi-class of objects in different lighting conditions from the Autonomous Excavator System (AES). We also evaluate our method on a general construction dataset and achieve SOTA results.

\bibliography{sn-bibliography}


\begin{thebibliography}{25}
\ifx \bisbn   \undefined \def \bisbn  #1{ISBN #1}\fi
\ifx \binits  \undefined \def \binits#1{#1}\fi
\ifx \bauthor  \undefined \def \bauthor#1{#1}\fi
\ifx \batitle  \undefined \def \batitle#1{#1}\fi
\ifx \bjtitle  \undefined \def \bjtitle#1{#1}\fi
\ifx \bvolume  \undefined \def \bvolume#1{\textbf{#1}}\fi
\ifx \byear  \undefined \def \byear#1{#1}\fi
\ifx \bissue  \undefined \def \bissue#1{#1}\fi
\ifx \bfpage  \undefined \def \bfpage#1{#1}\fi
\ifx \blpage  \undefined \def \blpage #1{#1}\fi
\ifx \burl  \undefined \def \burl#1{\textsf{#1}}\fi
\ifx \doiurl  \undefined \def \doiurl#1{\url{https://doi.org/#1}}\fi
\ifx \betal  \undefined \def \betal{\textit{et al.}}\fi
\ifx \binstitute  \undefined \def \binstitute#1{#1}\fi
\ifx \binstitutionaled  \undefined \def \binstitutionaled#1{#1}\fi
\ifx \bctitle  \undefined \def \bctitle#1{#1}\fi
\ifx \beditor  \undefined \def \beditor#1{#1}\fi
\ifx \bpublisher  \undefined \def \bpublisher#1{#1}\fi
\ifx \bbtitle  \undefined \def \bbtitle#1{#1}\fi
\ifx \bedition  \undefined \def \bedition#1{#1}\fi
\ifx \bseriesno  \undefined \def \bseriesno#1{#1}\fi
\ifx \blocation  \undefined \def \blocation#1{#1}\fi
\ifx \bsertitle  \undefined \def \bsertitle#1{#1}\fi
\ifx \bsnm \undefined \def \bsnm#1{#1}\fi
\ifx \bsuffix \undefined \def \bsuffix#1{#1}\fi
\ifx \bparticle \undefined \def \bparticle#1{#1}\fi
\ifx \barticle \undefined \def \barticle#1{#1}\fi
\bibcommenthead
\ifx \bconfdate \undefined \def \bconfdate #1{#1}\fi
\ifx \botherref \undefined \def \botherref #1{#1}\fi
\ifx \url \undefined \def \url#1{\textsf{#1}}\fi
\ifx \bchapter \undefined \def \bchapter#1{#1}\fi
\ifx \bbook \undefined \def \bbook#1{#1}\fi
\ifx \bcomment \undefined \def \bcomment#1{#1}\fi
\ifx \oauthor \undefined \def \oauthor#1{#1}\fi
\ifx \citeauthoryear \undefined \def \citeauthoryear#1{#1}\fi
\ifx \endbibitem  \undefined \def \endbibitem {}\fi
\ifx \bconflocation  \undefined \def \bconflocation#1{#1}\fi
\ifx \arxivurl  \undefined \def \arxivurl#1{\textsf{#1}}\fi
\csname PreBibitemsHook\endcsname

\bibitem{zhang2021autonomous}
\begin{botherref}
\oauthor{\bsnm{Zhang}, \binits{L.}},
\oauthor{\bsnm{Zhao}, \binits{J.}},
\oauthor{\bsnm{Long}, \binits{P.}},
\oauthor{\bsnm{Wang}, \binits{L.}},
\oauthor{\bsnm{Qian}, \binits{L.}},
\oauthor{\bsnm{Lu}, \binits{F.}},
\oauthor{\bsnm{Song}, \binits{X.}},
\oauthor{\bsnm{Manocha}, \binits{D.}}:
An autonomous excavator system for material loading tasks.
Science Robotics
\textbf{6}(55)
(2021)
\end{botherref}
\endbibitem

\bibitem{10.22260/ISARC2021/0009}
\begin{bchapter}
\bauthor{\bsnm{Zhang}, \binits{S.}},
\bauthor{\bsnm{Zhang}, \binits{L.}}:
\bctitle{Vision-based excavator activity analysis and safety monitoring
  system}.
In: \beditor{\bsnm{Feng}, \binits{C.}},
\beditor{\bsnm{Linner}, \binits{T.}},
\beditor{\bsnm{Brilakis}, \binits{I.}},
\beditor{\bsnm{Castro}, \binits{D.}},
\beditor{\bsnm{Chen}, \binits{P.-H.}},
\beditor{\bsnm{Cho}, \binits{Y.}},
\beditor{\bsnm{Du}, \binits{J.}},
\beditor{\bsnm{Ergan}, \binits{S.}},
\beditor{\bparticle{Garcia~de} \bsnm{Soto}, \binits{B.}},
\beditor{\bsnm{Ga~parík}, \binits{J.}},
\beditor{\bsnm{Habbal}, \binits{F.}},
\beditor{\bsnm{Hammad}, \binits{A.}},
\beditor{\bsnm{Iturralde}, \binits{K.}},
\beditor{\bsnm{Bock}, \binits{T.}},
\beditor{\bsnm{Kwon}, \binits{S.}},
\beditor{\bsnm{Lafhaj}, \binits{Z.}},
\beditor{\bsnm{Li}, \binits{N.}},
\beditor{\bsnm{Liang}, \binits{C.-J.}},
\beditor{\bsnm{Mantha}, \binits{B.}},
\beditor{\bsnm{Ng}, \binits{M.S.}},
\beditor{\bsnm{Hall}, \binits{D.}},
\beditor{\bsnm{Pan}, \binits{M.}},
\beditor{\bsnm{Pan}, \binits{W.}},
\beditor{\bsnm{Rahimian}, \binits{F.}},
\beditor{\bsnm{Raphael}, \binits{B.}},
\beditor{\bsnm{Sattineni}, \binits{A.}},
\beditor{\bsnm{Schlette}, \binits{C.}},
\beditor{\bsnm{Shabtai}, \binits{I.}},
\beditor{\bsnm{Shen}, \binits{X.}},
\beditor{\bsnm{Tang}, \binits{P.}},
\beditor{\bsnm{Teizer}, \binits{J.}},
\beditor{\bsnm{Turkan}, \binits{Y.}},
\beditor{\bsnm{Valero}, \binits{E.}},
\beditor{\bsnm{Zhu}, \binits{Z.}} (eds.)
\bbtitle{Proceedings of the 38th International Symposium on Automation and
  Robotics in Construction (ISARC)},
pp. \bfpage{49}--\blpage{56}.
\bpublisher{International Association for Automation and Robotics in
  Construction (IAARC)},
\blocation{Dubai, UAE}
(\byear{2021}).
\doiurl{10.22260/ISARC2021/0009}
\end{bchapter}
\endbibitem

\bibitem{wang2019predicting}
\begin{bchapter}
\bauthor{\bsnm{Wang}, \binits{M.}},
\bauthor{\bsnm{Wong}, \binits{P.}},
\bauthor{\bsnm{Luo}, \binits{H.}},
\bauthor{\bsnm{Kumar}, \binits{S.}},
\bauthor{\bsnm{Delhi}, \binits{V.}},
\bauthor{\bsnm{Cheng}, \binits{J.}}:
\bctitle{Predicting safety hazards among construction workers and equipment
  using computer vision and deep learning techniques}.
In: \bbtitle{ISARC. Proceedings of the International Symposium on Automation
  and Robotics in Construction},
vol. \bseriesno{36},
pp. \bfpage{399}--\blpage{406}
(\byear{2019}).
\bcomment{IAARC Publications}
\end{bchapter}
\endbibitem

\bibitem{ren2016faster}
\begin{barticle}
\bauthor{\bsnm{Ren}, \binits{S.}},
\bauthor{\bsnm{He}, \binits{K.}},
\bauthor{\bsnm{Girshick}, \binits{R.}},
\bauthor{\bsnm{Sun}, \binits{J.}}:
\batitle{Faster r-cnn: towards real-time object detection with region proposal
  networks}.
\bjtitle{IEEE transactions on pattern analysis and machine intelligence}
\bvolume{39}(\bissue{6}),
\bfpage{1137}--\blpage{1149}
(\byear{2016})
\end{barticle}
\endbibitem

\bibitem{redmon2018yolov3}
\begin{botherref}
\oauthor{\bsnm{Redmon}, \binits{J.}},
\oauthor{\bsnm{Farhadi}, \binits{A.}}:
YOLOv3: An Incremental Improvement
(2018)
\end{botherref}
\endbibitem

\bibitem{bochkovskiy2020yolov4}
\begin{botherref}
\oauthor{\bsnm{Bochkovskiy}, \binits{A.}},
\oauthor{\bsnm{Wang}, \binits{C.-Y.}},
\oauthor{\bsnm{Liao}, \binits{H.-Y.M.}}:
Yolov4: Optimal speed and accuracy of object detection.
arXiv preprint arXiv:2004.10934
(2020)
\end{botherref}
\endbibitem

\bibitem{glenn_jocher_2020_4154370}
\begin{botherref}
\oauthor{\bsnm{Jocher}, \binits{G.}},
\oauthor{\bsnm{Stoken}, \binits{A.}},
\oauthor{\bsnm{Borovec}, \binits{J.}},
\oauthor{\bsnm{NanoCode012}},
\oauthor{\bsnm{ChristopherSTAN}},
\oauthor{\bsnm{Changyu}, \binits{L.}},
\oauthor{\bsnm{Laughing}},
\oauthor{\bsnm{tkianai}},
\oauthor{\bsnm{Hogan}, \binits{A.}},
\oauthor{\bsnm{lorenzomammana}},
\oauthor{\bsnm{yxNONG}},
\oauthor{\bsnm{AlexWang1900}},
\oauthor{\bsnm{Diaconu}, \binits{L.}},
\oauthor{\bsnm{Marc}},
\oauthor{\bsnm{wanghaoyang0106}},
\oauthor{\bsnm{ml5ah}},
\oauthor{\bsnm{Doug}},
\oauthor{\bsnm{Ingham}, \binits{F.}},
\oauthor{\bsnm{Frederik}},
\oauthor{\bsnm{Guilhen}},
\oauthor{\bsnm{Hatovix}},
\oauthor{\bsnm{Poznanski}, \binits{J.}},
\oauthor{\bsnm{Fang}, \binits{J.}},
\oauthor{\bsnm{Yu}, \binits{L.}},
\oauthor{\bsnm{changyu98}},
\oauthor{\bsnm{Wang}, \binits{M.}},
\oauthor{\bsnm{Gupta}, \binits{N.}},
\oauthor{\bsnm{Akhtar}, \binits{O.}},
\oauthor{\bsnm{PetrDvoracek}},
\oauthor{\bsnm{Rai}, \binits{P.}}:
{ultralytics/yolov5: V3.1 - Bug Fixes and Performance Improvements}.
\doiurl{10.5281/zenodo.4154370}.
\url{https://doi.org/10.5281/zenodo.4154370}
\end{botherref}
\endbibitem

\bibitem{wang2020cspnet}
\begin{bchapter}
\bauthor{\bsnm{Wang}, \binits{C.-Y.}},
\bauthor{\bsnm{Liao}, \binits{H.-Y.M.}},
\bauthor{\bsnm{Wu}, \binits{Y.-H.}},
\bauthor{\bsnm{Chen}, \binits{P.-Y.}},
\bauthor{\bsnm{Hsieh}, \binits{J.-W.}},
\bauthor{\bsnm{Yeh}, \binits{I.-H.}}:
\bctitle{Cspnet: A new backbone that can enhance learning capability of cnn}.
In: \bbtitle{Proceedings of the IEEE/CVF Conference on Computer Vision and
  Pattern Recognition Workshops},
pp. \bfpage{390}--\blpage{391}
(\byear{2020})
\end{bchapter}
\endbibitem

\bibitem{raoofiamask}
\begin{botherref}
\oauthor{\bsnm{Raoofia}, \binits{H.}},
\oauthor{\bsnm{Motamedib}, \binits{A.}}:
Mask r-cnn deep learning-based approach to detect construction machinery on
  jobsites
\end{botherref}
\endbibitem

\bibitem{nakamura2020pose}
\begin{bchapter}
\bauthor{\bsnm{Nakamura}, \binits{H.}},
\bauthor{\bsnm{Tsukada}, \binits{Y.}},
\bauthor{\bsnm{Tamaki}, \binits{T.}},
\bauthor{\bsnm{Raytchev}, \binits{B.}},
\bauthor{\bsnm{Kaneda}, \binits{K.}}:
\bctitle{Pose estimation of excavators}.
In: \bbtitle{International Workshop on Advanced Imaging Technology (IWAIT)
  2020},
vol. \bseriesno{11515},
p. \bfpage{115152}
(\byear{2020}).
\bcomment{International Society for Optics and Photonics}
\end{bchapter}
\endbibitem

\bibitem{soltani2017skeleton}
\begin{barticle}
\bauthor{\bsnm{Soltani}, \binits{M.M.}},
\bauthor{\bsnm{Zhu}, \binits{Z.}},
\bauthor{\bsnm{Hammad}, \binits{A.}}:
\batitle{Skeleton estimation of excavator by detecting its parts}.
\bjtitle{Automation in Construction}
\bvolume{82},
\bfpage{1}--\blpage{15}
(\byear{2017})
\end{barticle}
\endbibitem

\bibitem{feichtenhofer2019slowfast}
\begin{bchapter}
\bauthor{\bsnm{Feichtenhofer}, \binits{C.}},
\bauthor{\bsnm{Fan}, \binits{H.}},
\bauthor{\bsnm{Malik}, \binits{J.}},
\bauthor{\bsnm{He}, \binits{K.}}:
\bctitle{Slowfast networks for video recognition}.
In: \bbtitle{Proceedings of the IEEE/CVF International Conference on Computer
  Vision},
pp. \bfpage{6202}--\blpage{6211}
(\byear{2019})
\end{bchapter}
\endbibitem

\bibitem{bertasius2021space}
\begin{botherref}
\oauthor{\bsnm{Bertasius}, \binits{G.}},
\oauthor{\bsnm{Wang}, \binits{H.}},
\oauthor{\bsnm{Torresani}, \binits{L.}}:
Is space-time attention all you need for video understanding?
arXiv preprint arXiv:2102.05095
(2021)
\end{botherref}
\endbibitem

\bibitem{ding2018deep}
\begin{barticle}
\bauthor{\bsnm{Ding}, \binits{L.}},
\bauthor{\bsnm{Fang}, \binits{W.}},
\bauthor{\bsnm{Luo}, \binits{H.}},
\bauthor{\bsnm{Love}, \binits{P.E.}},
\bauthor{\bsnm{Zhong}, \binits{B.}},
\bauthor{\bsnm{Ouyang}, \binits{X.}}:
\batitle{A deep hybrid learning model to detect unsafe behavior: Integrating
  convolution neural networks and long short-term memory}.
\bjtitle{Automation in construction}
\bvolume{86},
\bfpage{118}--\blpage{124}
(\byear{2018})
\end{barticle}
\endbibitem

\bibitem{chen2020automated}
\begin{barticle}
\bauthor{\bsnm{Chen}, \binits{C.}},
\bauthor{\bsnm{Zhu}, \binits{Z.}},
\bauthor{\bsnm{Hammad}, \binits{A.}}:
\batitle{Automated excavators activity recognition and productivity analysis
  from construction site surveillance videos}.
\bjtitle{Automation in construction}
\bvolume{110},
\bfpage{103045}
(\byear{2020})
\end{barticle}
\endbibitem

\bibitem{roberts2019end}
\begin{barticle}
\bauthor{\bsnm{Roberts}, \binits{D.}},
\bauthor{\bsnm{Golparvar-Fard}, \binits{M.}}:
\batitle{End-to-end vision-based detection, tracking and activity analysis of
  earthmoving equipment filmed at ground level}.
\bjtitle{Automation in Construction}
\bvolume{105},
\bfpage{102811}
(\byear{2019})
\end{barticle}
\endbibitem

\bibitem{bodla2017soft}
\begin{bchapter}
\bauthor{\bsnm{Bodla}, \binits{N.}},
\bauthor{\bsnm{Singh}, \binits{B.}},
\bauthor{\bsnm{Chellappa}, \binits{R.}},
\bauthor{\bsnm{Davis}, \binits{L.S.}}:
\bctitle{Soft-nms--improving object detection with one line of code}.
In: \bbtitle{Proceedings of the IEEE International Conference on Computer
  Vision},
pp. \bfpage{5561}--\blpage{5569}
(\byear{2017})
\end{bchapter}
\endbibitem

\bibitem{redmon2016you}
\begin{bchapter}
\bauthor{\bsnm{Redmon}, \binits{J.}},
\bauthor{\bsnm{Divvala}, \binits{S.}},
\bauthor{\bsnm{Girshick}, \binits{R.}},
\bauthor{\bsnm{Farhadi}, \binits{A.}}:
\bctitle{You only look once: Unified, real-time object detection}.
In: \bbtitle{Proceedings of the IEEE Conference on Computer Vision and Pattern
  Recognition},
pp. \bfpage{779}--\blpage{788}
(\byear{2016})
\end{bchapter}
\endbibitem

\bibitem{xiao2018simple}
\begin{bchapter}
\bauthor{\bsnm{Xiao}, \binits{B.}},
\bauthor{\bsnm{Wu}, \binits{H.}},
\bauthor{\bsnm{Wei}, \binits{Y.}}:
\bctitle{Simple baselines for human pose estimation and tracking}.
In: \bbtitle{Proceedings of the European Conference on Computer Vision (ECCV)},
pp. \bfpage{466}--\blpage{481}
(\byear{2018})
\end{bchapter}
\endbibitem

\bibitem{he2016deep}
\begin{bchapter}
\bauthor{\bsnm{He}, \binits{K.}},
\bauthor{\bsnm{Zhang}, \binits{X.}},
\bauthor{\bsnm{Ren}, \binits{S.}},
\bauthor{\bsnm{Sun}, \binits{J.}}:
\bctitle{Deep residual learning for image recognition}.
In: \bbtitle{Proceedings of the IEEE Conference on Computer Vision and Pattern
  Recognition},
pp. \bfpage{770}--\blpage{778}
(\byear{2016})
\end{bchapter}
\endbibitem

\bibitem{lin2014microsoft}
\begin{bchapter}
\bauthor{\bsnm{Lin}, \binits{T.-Y.}},
\bauthor{\bsnm{Maire}, \binits{M.}},
\bauthor{\bsnm{Belongie}, \binits{S.}},
\bauthor{\bsnm{Hays}, \binits{J.}},
\bauthor{\bsnm{Perona}, \binits{P.}},
\bauthor{\bsnm{Ramanan}, \binits{D.}},
\bauthor{\bsnm{Doll{\'a}r}, \binits{P.}},
\bauthor{\bsnm{Zitnick}, \binits{C.L.}}:
\bctitle{Microsoft coco: Common objects in context}.
In: \bbtitle{European Conference on Computer Vision},
pp. \bfpage{740}--\blpage{755}
(\byear{2014}).
\bcomment{Springer}
\end{bchapter}
\endbibitem

\bibitem{chen2019mmdetection}
\begin{botherref}
\oauthor{\bsnm{Chen}, \binits{K.}},
\oauthor{\bsnm{Wang}, \binits{J.}},
\oauthor{\bsnm{Pang}, \binits{J.}},
\oauthor{\bsnm{Cao}, \binits{Y.}},
\oauthor{\bsnm{Xiong}, \binits{Y.}},
\oauthor{\bsnm{Li}, \binits{X.}},
\oauthor{\bsnm{Sun}, \binits{S.}},
\oauthor{\bsnm{Feng}, \binits{W.}},
\oauthor{\bsnm{Liu}, \binits{Z.}},
\oauthor{\bsnm{Xu}, \binits{J.}}, et al.:
Mmdetection: Open mmlab detection toolbox and benchmark.
arXiv preprint arXiv:1906.07155
(2019)
\end{botherref}
\endbibitem

\bibitem{mmseg2020}
\begin{botherref}
\oauthor{\bsnm{Contributors}, \binits{M.}}:
{MMSegmentation}: OpenMMLab Semantic Segmentation Toolbox and Benchmark.
\url{https://github.com/open-mmlab/mmsegmentation}
(2020)
\end{botherref}
\endbibitem

\bibitem{mmpose2020}
\begin{botherref}
\oauthor{\bsnm{Contributors}, \binits{M.}}:
OpenMMLab Pose Estimation Toolbox and Benchmark.
\url{https://github.com/open-mmlab/mmpose}
(2020)
\end{botherref}
\endbibitem

\bibitem{2020mmaction2}
\begin{botherref}
\oauthor{\bsnm{Contributors}, \binits{M.}}:
OpenMMLab's Next Generation Video Understanding Toolbox and Benchmark.
\url{https://github.com/open-mmlab/mmaction2}
(2020)
\end{botherref}
\endbibitem

\end{thebibliography}


\end{document}